\documentclass[sigconf,screen,nonacm]{acmart}

\AtBeginDocument{%
  \providecommand\BibTeX{{%
    \normalfont B\kern-0.5em{\scshape i\kern-0.25em b}\kern-0.8em\TeX}}}

\setcopyright{none}
\copyrightyear{2022}
\acmYear{2022}
\acmDOI{XXXXXXX.XXXXXXX}

\usepackage{multirow}
\usepackage{xcolor}

\usepackage{comment}
\usepackage{array}
\usepackage{titlesec}
\usepackage{amsmath,mathtools}
\usepackage{graphicx} 
\usepackage{subcaption}
\usepackage{caption}
\usepackage{bm}
\usepackage{tabularray}
\usepackage{soul}
\usepackage{capt-of,etoolbox}
\usepackage{graphicx}
\graphicspath{ {./figs/} }
\usepackage{xspace}
\usepackage{tabularx}

\usepackage[bottom]{footmisc}

\usepackage{hyperref}
\title[VR Isle Academy: A VR Digital Surgical Robotic System]{ VR Isle Academy: A VR Digital Twin Approach for Robotic Surgical Skill Development}

\begin{document}

\author{Achilleas Filippidis}
\orcid{0009-0007-2826-7494}
\email{achilles.filippidis@oramavr.com}
\affiliation{%
  \institution{FORTH - ICS, University of Crete, ORamaVR}
  \city{Heraklion}
  \country{Greece}
}

\author{Nikolaos Marmaras}
\orcid{0009-0007-1714-4026}
\email{n.marmaras@hotmail.com}
\affiliation{%
  \institution{University of Western Macedonia, ORamaVR}
  \city{Kozani}
  \country{Greece}
}

\author{Michael Maravgakis}
\orcid{0009-0009-2345-9264}
\email{maravgakis@ics.forth.gr}
\affiliation{%
  \institution{FORTH - ICS, University of Crete, ORamaVR}
  \city{Heraklion}
  \country{Greece}
}

\author{Alexandra Plexousaki}
\orcid{0009-0009-2909-4753}
\email{aplex@ics.forth.gr}
\affiliation{%
  \institution{FORTH - ICS, University of Crete, ORamaVR}
  \city{Heraklion}
  \country{Greece}
}

\author{Manos Kamarianakis}
\email{kamarianakis@uoc.gr}
\orcid{0000-0001-6577-0354}
\affiliation{%
  \institution{FORTH - ICS, University of Crete, ORamaVR}
  \city{Heraklion}
  \country{Greece}
}

\author{George Papagiannakis}
\orcid{0000-0002-2977-9850}
\email{papagian@ics.forth.gr}
\affiliation{%
  \institution{FORTH - ICS, University of Crete, ORamaVR}
  \city{Heraklion}
  \country{Greece}
}

\renewcommand{\shortauthors}{Filippidis et al.}


\begin{teaserfigure}
    \centering
    \includegraphics[width=\textwidth]{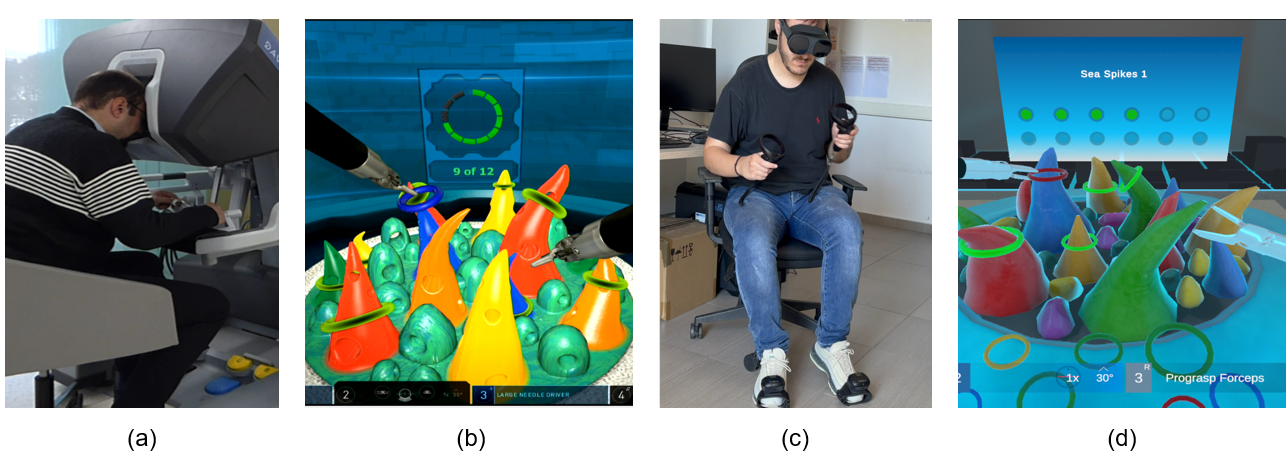}
    \caption{An actual Surgical Robotic System (SRS) simulation in comparison with its digital twin, VR Isle Academy. In images (a) and (b), a modern SRS simulator is depicted, showing a user operating from the surgeon's console. In contrast, images (c) and (d) showcase VR Isle Academy, where the user controls a simulated SRS digital-twin using an inside-out VR HMD, controllers and feet trackers.}
    \label{fig:myteaser}
\end{teaserfigure}

\begin{abstract} % Todo: cannot be seen.

Contemporary progress in the field of robo\-tics, marked by improved
efficiency and stability, has paved the way for the global adoption of
surgical robotic systems (SRS).
While these systems enhance surgeons' skills by offering a more
accurate and less invasive approach to operations, they come at a
considerable cost.
Moreover, SRS components often involve heavy machinery, making the
training process challenging due to limited access to such equipment.
In this paper we introduce a cost-effective way to facilitate training
for a simulator of a SRS via a portable,
device-agnostic, ultra realistic simulation with hand tracking and
feet tracking support. 
Error assessment is accessible in both real-time and offline,
which enables the monitoring and tracking of users' performance. 
The VR application has been objectively evaluated by several untrained
testers showcasing significant reduction in error metrics as the
number of training sessions increases. 
This indicates that the proposed VR application denoted as
\textit{VR Isle Academy} operates efficiently, improving the robot - controlling
skills of the testers in an intuitive and immersive way towards
reducing the learning curve at minimal cost.

\end{abstract}

\settopmatter{printfolios=true}
\maketitle

\section{Introduction}

In recent years, the trajectory of medical training has significantly
shifted, incorporating the latest technological advancements. 
However, this transition is not extensively adopted in universities or
other institutions that focus on medical training, primarily due to
the limited availability of market products and the high cost of the
required training equipment. To this day, due to the aforementioned
reasons, the majority of surgical training courses still adhere to a
pattern of the 16th-century training procedure~\cite{medicineHistory}
where the trainees simply observe the expert-surgeon/tutor perform
surgery. 

Contemporary advancements in the field of robotics have established
robotic surgical systems as a viable option for performing highly
precise minimally invasive operations enabling the surgeon to operate
while seated. Some of the surgical robotic systems that are out in the
market are the da Vinci surgical
system (\url{https://www.intuitive.com/en-us/products-and-services/
da-vinci}), Senhance surgical system (\url{https://www.asensus.com/})
and Flex robotic
system (\url{https://novusarge.com/en/medical-products/flex-robotic
-system/}). The da Vinci Surgical System \cite{davinci_robot} has is one of the most widely used robotic surgical systems \cite{reviewOfSRS}. This system has been used for many
different operations such as cardiac, colorectal, general,
gynecologic, head and neck, thoracic, and urologic
surgeries~\cite{reviewOfSRS}. In 2021, 6500 da Vinci Surgical system were installed in 67 different countries and 55.000 doctors were trained to use it~\cite{intuitive_info_about_machines_worldwide,davinci_stats}.
The cost of acquiring and maintaining the above Surgical System is significant. Due to cost considerations of acquiring it and
the low amount of systems around the world, various companies have capitalized on private training courses tailored for doctors and surgeons.

The field of Virtual Reality (VR) has undergone major advancements
with powerful VR headsets being able to render entire worlds in
real-time. This has introduced a new market for VR, in medical
training. VR training offers an immersive experience for the trainees
who enhance their hard-skills inside the virtual world
and gain experience by training their hard-skills. 
Researchers across the globe have directed their efforts towards
enhancing the scientific domain of VR medical training by introducing
innovative solutions to address existing challenges, such as those
highlighted in \cite{supernumeraryroboticlimbs},
\cite{Intracardiac_VR}.

Recognizing the necessity for a more convenient, affordable, and
portable approach to utilize SRS, we
suggest an advanced VR Ultra Realistic training simulation for
surgical robotic systems. Figure~\ref{fig:myteaser}a illustrates the
user utilizing the machine that controls the robotic arms, with
figure~\ref{fig:myteaser}b depicting the view from the simulated
training scenario. Figures~\ref{fig:myteaser}c and
~\ref{fig:myteaser}d respectively demonstrate a user being trained in
the same scenario using our application along with his view within VR.
This VR simulation democratizes the training
of these systems with a ``device-agnostic'' strategy by reducing the
cost of training and smoothing out the learning curve. The
incorporation of feet tracking enhances user
immersion, providing an authentic training experience for a surgical
robotic system. 

The main contribution of this work is to present a complete digital-twin
of the SRS training process. To the best of our knowledge,
VR Isle Academy is the first approach that provides
the full training experience entirely in VR. In this
paper, we selected da Vinci as the reference point due to its renowned
reputation within the global community of surgeons. However, the work
accomplished can be adapted to replicate any SRS system and not just
the mechanics and training scenarios of the reference SRS. By
leveraging available tools, we've developed a VR simulation enabling
trainees to undergo SRS training conveniently, irrespective of location
or time constraints. This addresses the challenge posed by the limited
availability of SRS training devices in certain geographical areas,
thereby saving both time and expenses associated with traditional
training methods.

\section{Related Work}
\subsection{Digital Twin} 
A Digital Twin is a virtual representation, mirroring a physical
object or process in the digital realm with a high-fidelity
resemblance. The term was publicly introduced by Michael Grieves for a
product lifecycle management~\cite{explanation_of_digitalTwin}.

In the modern age, digital twins are extensively utilized across
various sectors including power generation equipment, structures,
manufacturing operations, automotive industry, healthcare services,
and urban planning~\cite{IBM}. Specifically, in domains such as SRS
training, digital twins offer opportunities to simulate either
real-life procedures, like laparoscopic surgery using the SRS, or
typical training scenarios utilized for doctor training in SRS
procedures. Within the framework of our project, we've developed a
digital twin of the SRS training simulator, which can simulate real
surgical operations when necessary.

\subsection{Medical Training in VR}  

Numerous efforts have been made to expedite the training and education process in the medical field using VR. Recently, the cost of acquiring and maintaining commercial VR head-mounted displays (HMDs) has decreased. Furthermore, the contemporary advancements in HMD technology significantly enhance the overall performance of standalone applications. To this end, VR technology has been widely adopted for facilitating medical training not only for students but also for health care professionals. Several research papers and examples have demonstrated that VR training in the medical field reduces malpractices, training time, and the learning curve~\cite{orthopedical_surgery_ossoVR,tibial_shaft_ossoVR,paper_from_fundamental,Kenanidis2023}. To facilitate the development of medical training scenarios, several software development kits (SDK) have been released.

MAGES SDK~\cite{mages4_0} is an innovative SDK that empowers developers with numerous tools to efficiently create fast and effective medical training scenarios. Paul Zikas \textit{et al.} \cite{mages4_0} highlight the latest advancements in the aforementioned SDK, including 5G edge-cloud remote rendering, a physics dissection layer, real-time simulation of organic tissues as soft-bodies within 10 ms, a highly realistic cutting and tearing algorithm, neural network assessment for user profiling, and a VR recorder for recording, replaying, and debriefing training simulations from any perspective. 

Fundamental Core~\cite{fundamental_core} is an all-in-one SDK for Unity game engine. The developer has the capability to establish a scoring system for real-time results at the end of a playthrough. Additionally, they provide a ready-to-use multiplayer service enabling users to connect and train together, complete with voice communication. Lastly, the SDK is device-agnostic and compatible with various VR headsets.

\subsection{The da Vinci Surgical Robotic System}

The da Vinci machine, a Surgical Robotic System developed by
Intuitive (\url{https://intuitive.com}), stands as the most
widely utilized SRS globally. This surgical system provides
the surgeon with an advanced set of instruments for conducting robotic-assisted minimally invasive surgery. It consists of a surgeon's console, the four robotic arms that are scissors, scalpel, 3D cameras and forceps that are connected and moved from the surgeon's console and the vision cart which makes the connection between the surgeon's console and the robotic arms. 

Moreover, the surgeon is provided with a superior vision, through the
3D real-time high-definition view with a magnifier that can reach up
to 10 times more than the human eye can see. Moreover, the ergonomic
design of the surgeon's console allows the surgeon to operate while
seated for extended periods, ensuring high efficiency in incisions.
The design also provides the surgeon with the capability to utilize
hand controllers and foot pedals for the machine's various
functionalities. 

Lastly, the machine offers various functionalities triggered by
pedals, masters, or the touch screen. The Camera Pedal allows
adjustment of the position and orientation of the camera attached to a
robotic arm. The Clutch Pedal is used to extend or shorten the robotic
arm. Four energy pedals control the electro-surgical instruments. The
30-degree view pedal toggles between different camera views.

\begin{figure}
\begin{center}
\includegraphics[width=1\columnwidth]{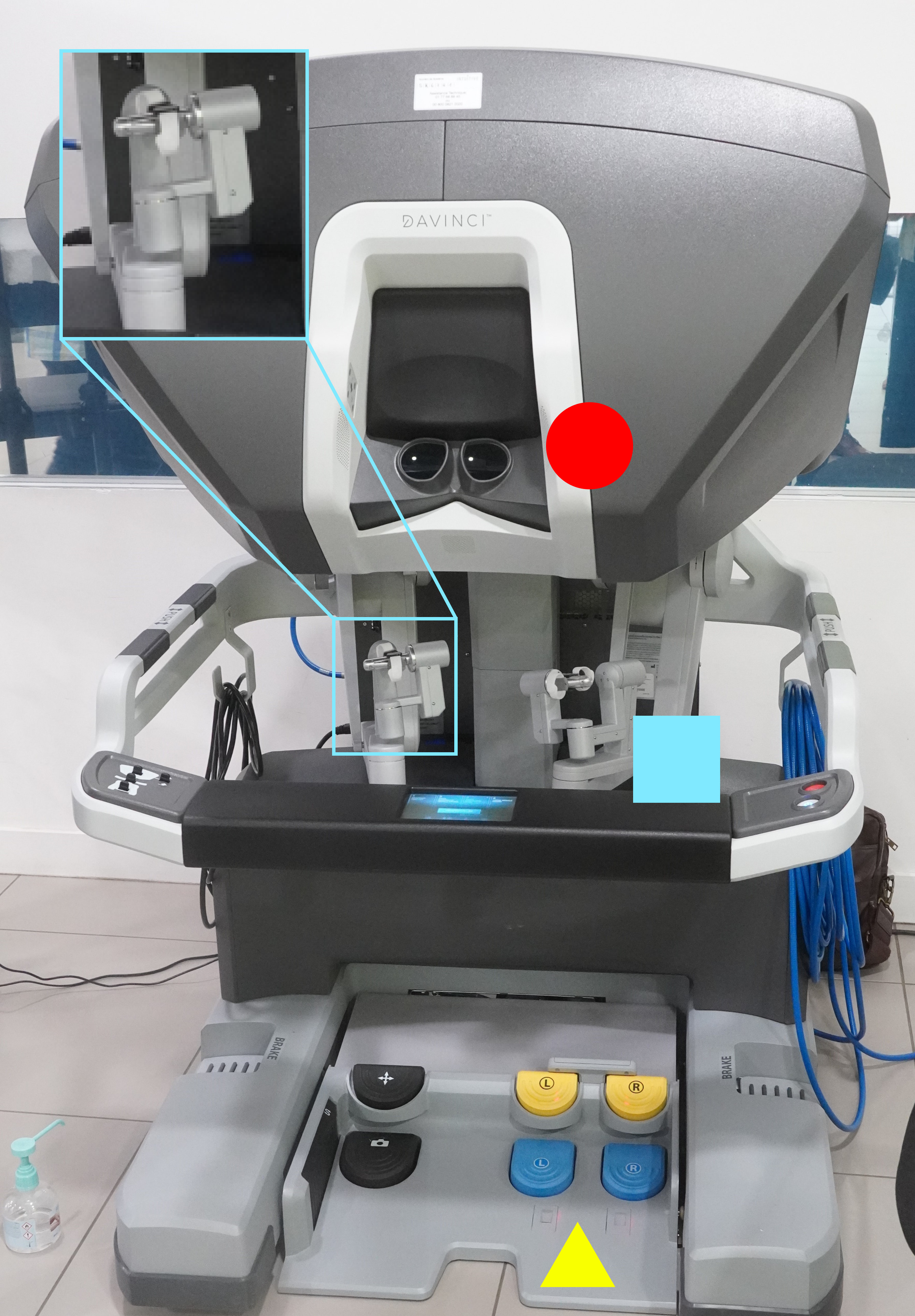}
\end{center}
\caption{The da Vinci surgeon's console features distinct elements: the yellow triangle represents the machine's pedals, allowing users to enable/disable various robot functionalities. The light blue square denotes the masters, where the surgeon controls the robotic arms. Each master typically features two rings, into which the surgeon places their fingers in order to control the rotation and movement of the robotic arms. Lastly, the red circle indicates the output of the cameras. Figure from \cite{davinci_wikipedia}} 
\label{fig:davinci_surgeon_console}
\end{figure}

\subsection{Previous Work}

The current bibliography includes numerous VR training
simulations incorporating advanced cognitive and psychomotor
techniques aimed at maximizing educational advantages for trainees,
exemplified by references such as \cite{oramavr_cvrsb} and
\cite{Kenanidis2023}. However, despite this proliferation, simulations
tailored for training in SRS within XR environments remain relatively
scarce.

Sketchy Neurons (\url{https://sketchyneurons.com/}) have
created a VR game called \textit{Minimally
Invasive} (\url{https://store.steampowered.com/app/2331420/Minimally_Invasive/}) that uses a surgical robotic system. The concept of
the game is that the user is a transplant surgeon and will have to
burn, slice and operate aliens. They claim to have realistic physics
and that the tools that are used are developed by actual surgeons. In
the VR game, you can train on how to use and operate the robotic
arms. They have created a menu where you can select to train and
learn how to move and operate the robotic arms. The user will encounter
six scenarios designed to teach functionalities similar to those of a
clutch and camera.

\balance
While they have embraced a device agnostic approach by incorporating SteamVR (\url{https://store.steampowered.com/app/250820/SteamVR/}), it's worth noting that the game requires tethering, which significantly reduces po\-rta\-bility. Moreover, the application lacks realism by not imposing hand restrictions on the user, as opposed to a real SRS. Lastly, the physics appears to be implemented in a manner that lacks realism. 

Surgical Robot Simulator (\url{https://store.steampowered.com/app/1727070/Surgical_Robot_Simulator/}) is another VR serious game that incorporates the fundamentals of an SRS. The game provides tutorials on controller usage and offers a range of scenarios to engage with. A notable feature of this game is the utilization of deformation algorithms, allowing users to cut and manipulate deformable meshes. Despite its realistic graphics and the mesh deformation algorithm, the control scheme for manipulating forceps and robotic arms is neither optimal nor intuitive. Users can only control the movement and rotation of the robotic arms, while the rotation of the forceps is adjusted via the thumbstick. This configuration makes it challenging to intuitively and seamlessly manipulate the medical tool of the robot.

Xiaoyu Cai \textit{et al.} \cite{digitalTwin_CaiEtAl} proposed a
robotic minimally invasive surgical simulator based on VR digital. In
their research they used Pimax (\url{https://pimax.com/}) for the VR headset and two
3DSystems (\url{https://www.3dsystems.com/haptics-devices/touch}) Touch devices and two UR5 robots (\url{https://www.universal-robots.com/}). While they have successfully linked the
virtual and physical realms, there are still certain elements they are
missing. 
Primarily, the application lacks portability, as it necessitates the
presence of the robotic arm, touch devices, and the large machine
designed to simulate the SRS. Furthermore, they do not incorporate any
pedals, whether virtual or physical, to activate essential functions
such as the clutch and camera. Finally, their solution is not cost-effective, mainly due to the necessary equipment required for
the system to function effectively.

Marco Ferro \textit{et al.} \cite{ferro2019_portabeDavinci} proposed a
portable da Vinci simulator in VR using cheap haptic interfaces and an
Oculus Rift (\url{https://www.oculus.com}) to replicate the master console of the da Vinci. Despite its affordability, the system
lacks portability due to the use of two styluses and a tethered
connection to a desktop PC. Additionally, immersion is constrained,
with users confined to a training scene, while clutch functionality is implemented through stylus' buttons.

\section{A digital twin for the surgical robotic system}

\subsection{The Unity Game Engine and the MAGES SDK}

Unity (\url{https://unity.com/}) is a cross-platform game
engine that can be used to create two-dimensional and
three-dimensional games. The engine offers a primary scripting API in
C\#. We used a variety of external plugins in order to create
this digital twin environment. 
The core plugin we used in order to create each training scenario is
MAGES SDK, a robust tool enabling
the creation of immersive XR simulations. From this kit, we used the
analytics engine in order to capture the events needed to provide the
user with realtime and offline feedback, and scores depending on their performance.

Also, the VR editor facilitates the quick and straightforward development
of scenarios. Using scenegraph, a virtual editor that lets you create action nodes and modify existing ones in order to form a training scenario, we were able to create some of the basic scenarios of a SRS. More about them in the section \ref{Scenarios}.

Action nodes correspond to a certain task that has to be completed in
VR. The developer can either use one of the predefined action types,
such as insert action, remove action, use action, or create his own
action type.

Although our implementation is Unity-based, our approach can be
leveraged and implemented into any modern game engine.

\subsection{Robot Description} \label{Robot Description}
\begin{figure}
    \centering
    \includegraphics[width=0.9\columnwidth]{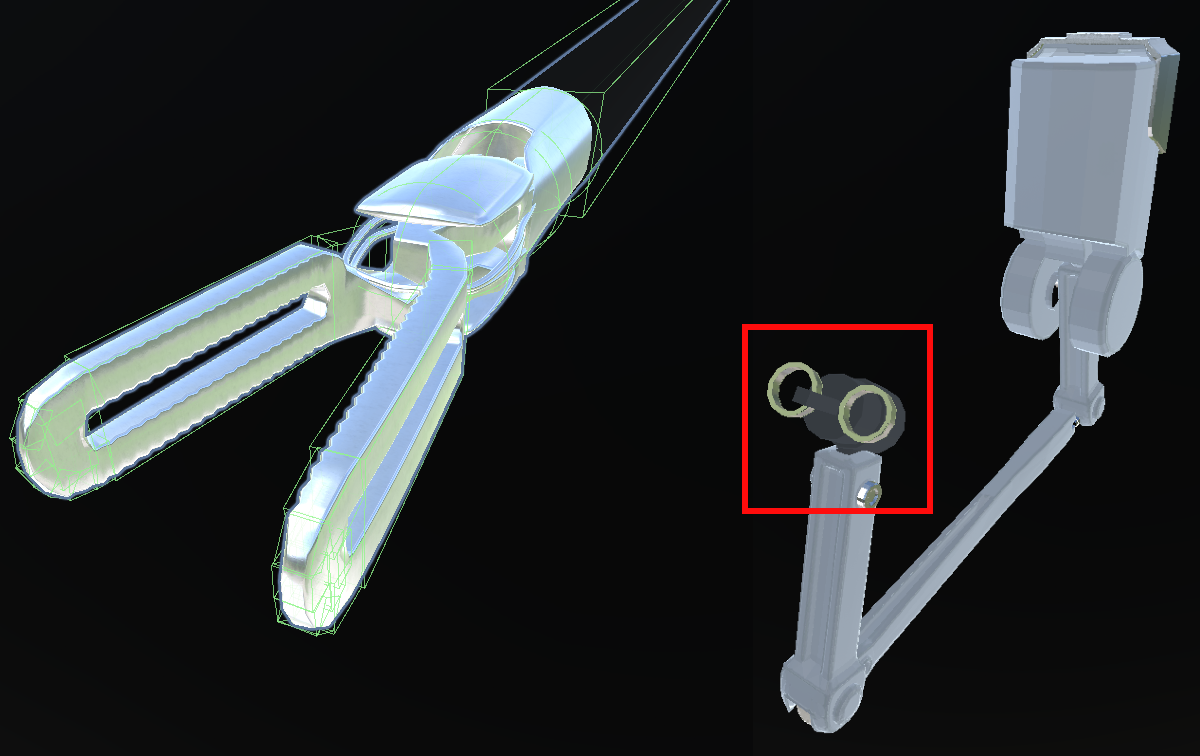}
    \caption{The digital-twin robotic arm's end effector (left) and the master control of the machine (right). The master controls the end effector in the digital-twin VR Isle Academy. The red box highlights where the doctor's fingers
    should be placed in the master during the operation.}
    \label{fig:master_slave_fig}
\end{figure}

The surgical robotic system employed in the VR environment consists of
two parts, which are commonly coined as \textit{master and
slave}~\cite{master_slave}. 
The master platform is controlled by the user with two-joystick-like
controls (Fig.~\ref{fig:master_slave_fig}) and two pedals that can be utilized for changing
functionalities when necessary.
On the other hand, the slave is located in a different area within the
scene and consists of two 6-Degrees of Freedom (DoFs) robotic arms. 
Each robotic arm is equipped with a 1-DoF two-jaw gripper end effector (the tool mounted at the end of the robotic arm)
featuring multiple box collider (Fig.~\ref{fig:master_slave_fig}), enhancing the realism and physics
accuracy of haptic interactions with objects of various shapes within
the scene. 
The master platform was acquired from~\cite{operator_platform} and it
was modified appropriately to improve rendering speed and the overall
efficiency of the model. 
The robotic arm was designed entirely by our team while drawing inspiration
from the specifications of the da Vinci surgical
robot (\url{https://www.intuitive.com/en-us/products-and-services/da-vinci}), the most widely utilized robotic system 
for minimally invasive surgeries~\cite{davinci_stats}.
Moreover, we aimed to replicate the surgeon's console, allowing users
to customize the head height of the machine and adjust the position of
the pedals to optimize ergonomic posture. Users can also re-calibrate
the trackers to easily configure their height. These operations can be
performed conveniently through the virtual tablet on the console.

\subsection{Training Scenarios} \label{Scenarios}

In our application, we simulated eleven scenarios basic resembling those found in a modern SRS. These scenarios are designed to familiarize the user with the control of the robotic arms, the clutch pedal, the camera pedal, and the 30-degree camera.
The purpose of these activities is to provide the user with an
interactive introduction to the surgical system's features. More
precisely, some exercises try to mix various functionalities in a
single scenario, while other lessons concentrate on a single one, such
as the camera function in the Camera 0 scenario. For example, in the
Sea Spikes exercises the user must learn how to manage delicate wrist
movements while combining the camera and clutch functionalities
effectively.

\begin{figure}
    \centering
    \includegraphics[width=0.9\columnwidth]{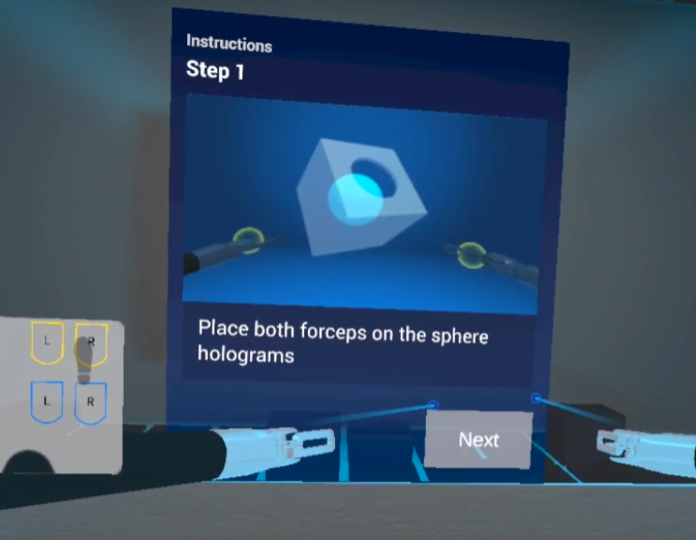}
    \caption{Instructions for the Wrist Articulation 1 exercise. Each exercise includes instructional steps to guide the user, explaining their objectives and detailing what actions to take and what to avoid.}
    \label{fig:tutorial_steps_1}
\end{figure}
 
\begin{figure}[h]
    \centering
    \includegraphics[width=0.9\columnwidth]{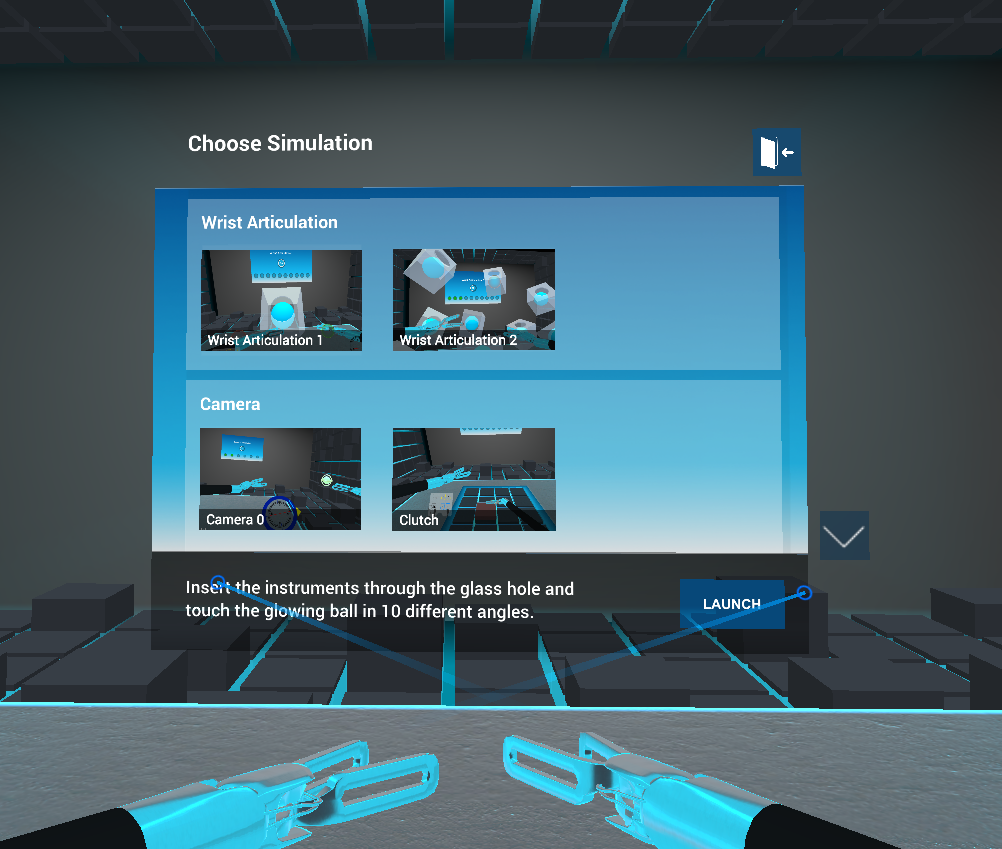}
    \caption{The exercise selection UI in the digital-twin. The user can select from a total of 12 scenarios for training purposes.}
    \label{fig:level_Selection}
\end{figure}

Through a User-Interface (UI) menu (Fig. \ref{fig:level_Selection}),
the users can choose to train in any one of the 12 scenarios.
Instructions (Fig. \ref{fig:tutorial_steps_1}) detailing the
objective of the exercise and the user's responsibilities are provided
for each scenario. For instance, in Wrist Articulation 1, users are
instructed to touch the ball inside the glass cube, without breaking
the exterior glass.

\begin{figure}
    \centering
    \includegraphics[width=0.9\columnwidth]{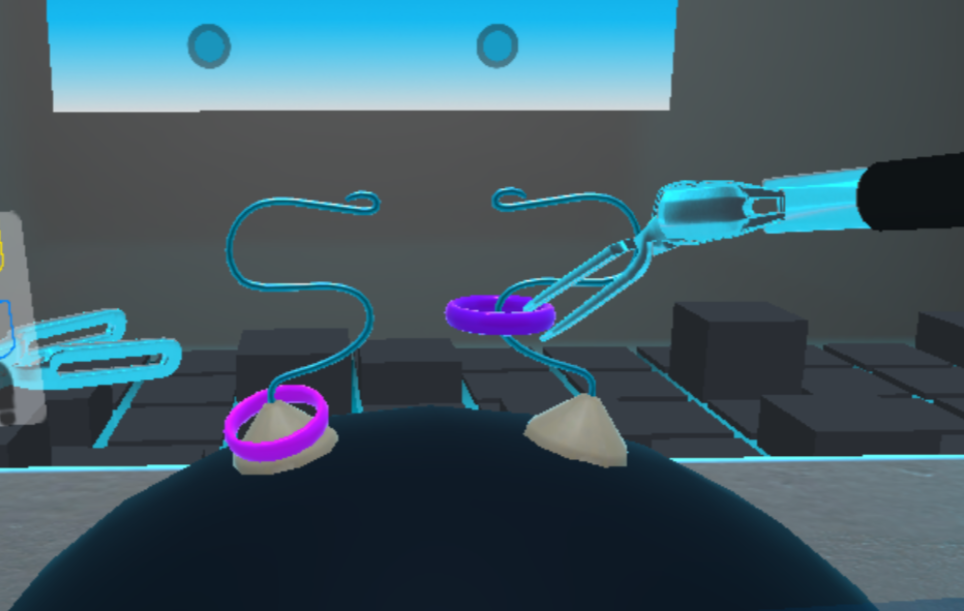}
    \caption{Ring Tower Transfer 1 scenario. In this scenario, the user is tasked with removing a ring from a wire tower and placing it in a specific position on the sides of spherical objects. Applying excessive force with the forceps or moving too quickly can result in the tower detaching, leading to a failure.}
    \label{fig:scenario_1}
\end{figure}

\begin{figure}
    \centering
    \includegraphics[width=0.9\columnwidth]{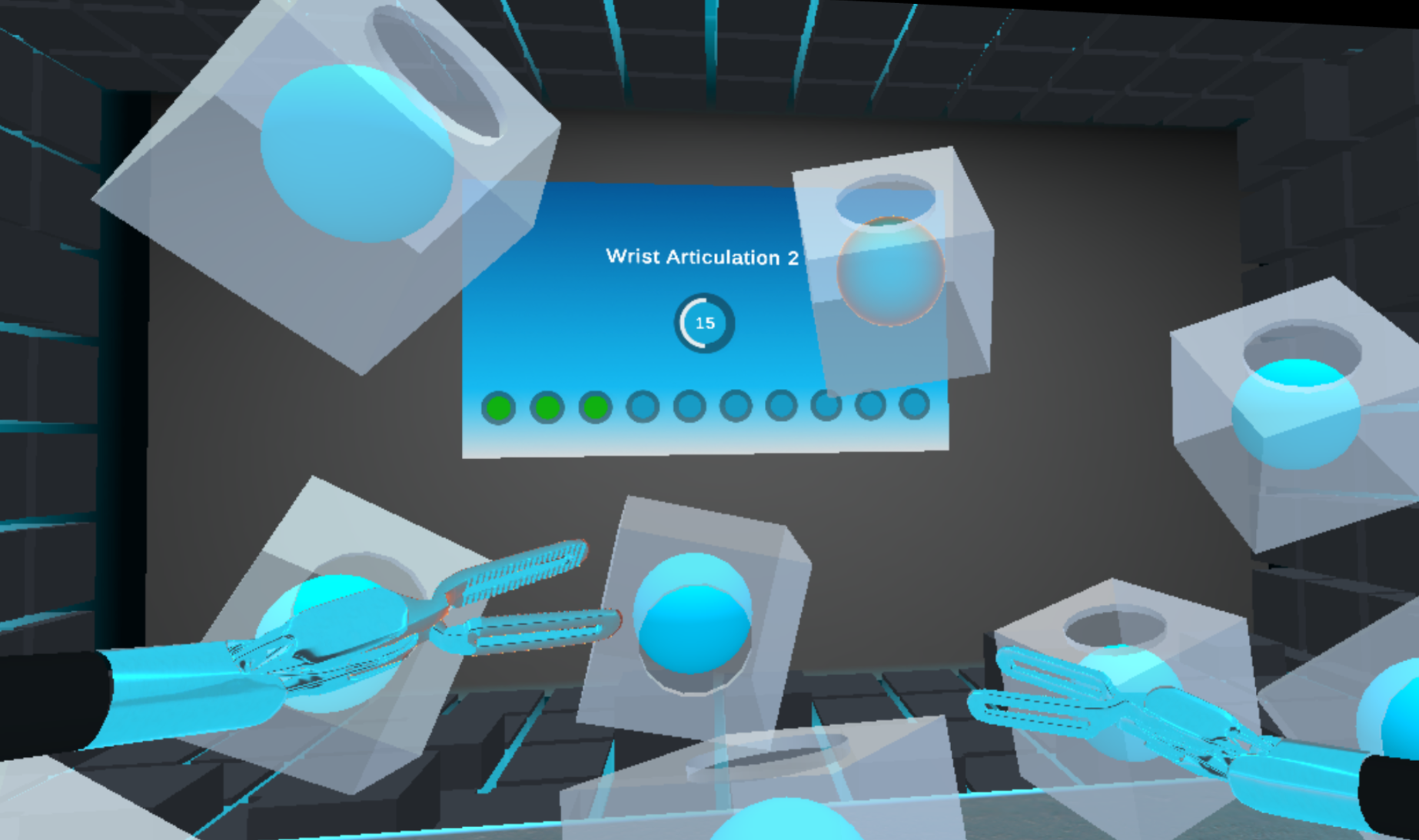}
    \caption{The Wrist Articulation 2 scenario in VR Isle Academy. The user is required to use the camera pedal to locate the correct ball. Subsequently, utilizing the clutch functionality, they must extend the robotic arms and, with precise wrist manipulations, touch the correct ball.}
    \label{fig:scenario_2}
\end{figure}

The implementation of the exercises was carried out using the scenegraph
framework from the MAGES SDK. In
this framework, exercises can be seen as Actions that users have to
perform. Exercises that require repetition, such as Wrist Articulation
1 or Camera 0, are implemented as one action that is repeated $X$
times, where $X$ is the amount of total iterations the scenario
requires. To exemplify, in Wrist Articulation 1, the users have to
perform two actions (in this case, place the instruments on a specific
position and touch the glowing ball) ten times, but on a different
angle. Lastly, more scenarios can be seamlessly added to the application in order to enhance the variety of the training options.

\subsection{Error Detection and Analytics}

In this section, we elucidate the scoring system we include in our
application, used for the qualitative assessment of the users'
performance. Through error detection and analytic metrics for each
exercise, the users can monitor their progress and improve their
skills. Each exercise contains a list of efficiency and penalty scores
used to assess the overall score of the session. The scoring factors
and analytics metrics of VR Isle Academy were designed and implemented
based on the corresponding factors of a modern SRS, 
with a focus on retaining the different weight and importance of each metric to the final score of each exercise.

\begin{figure}
\centering
\includegraphics[width=0.48\textwidth]{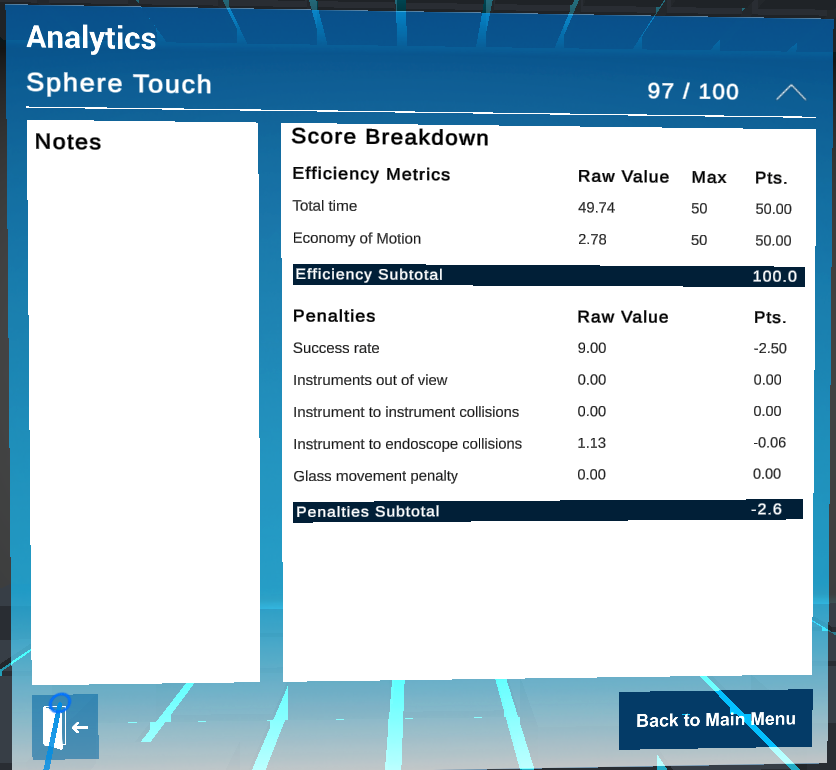}\\
\includegraphics[width=0.48\textwidth]{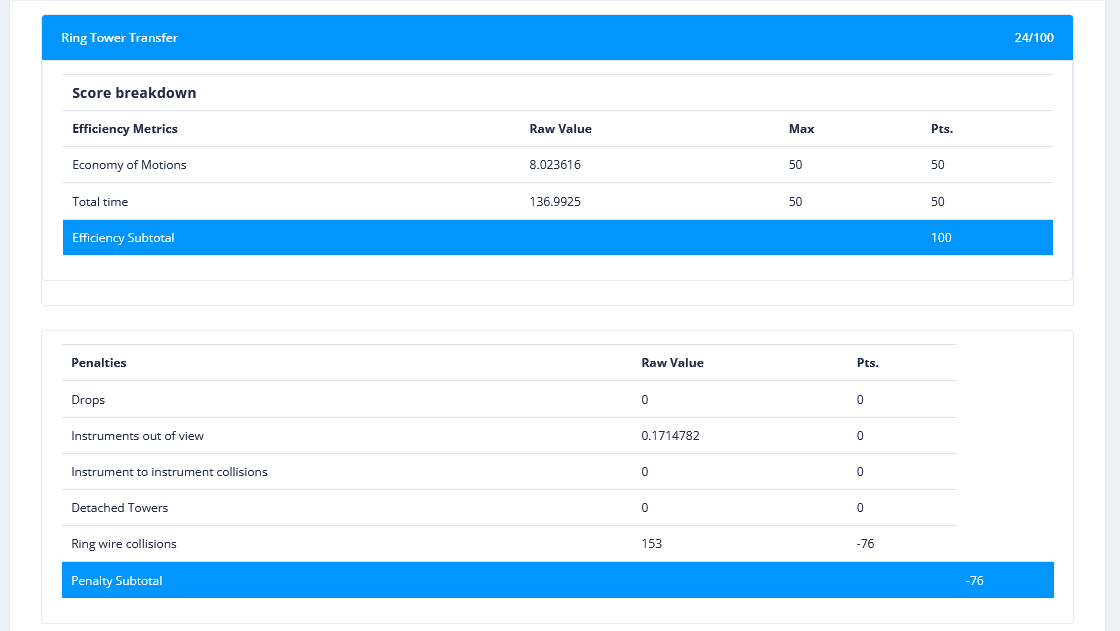}
\caption{The analytic metrics score breakdown. (Top) The analytics UI in the digital-twin. (Bottom) The analytics UI in the browser Portal.}
\label{fig:analyticsVR}
\end{figure}

The main focus of VR Isle Academy regarding the scoring system is to
retain the importance of each metric when providing user performance
feedback. In order to extract information regarding the scoring factors
and their importance, an iterative procedure was followed, where we
carefully examined the scores and their breakdown in the actual
simulation for each and every exercise. Each metric score is
calculated using a weight factor. This weight factor varies from metric to metric. For instance, the "drops" metric
which represents the times an object
held by the instruments fell on the environment, has a bigger weight
factor than the "excessive force" metric, which measures the times
excessive force was applied by the instruments to an object, while the
"tower detach" metric fails the exercise immediately.

The total score of an action is formed as such: \\
\begin{equation*}
\text{Score} := \sum \text{(Efficiency Metrics)} - \sum \text{(Penalty Scores)} \\
\end{equation*}

As can be observed, the analytic metrics are split into two main categories:
Efficiency Metrics and Penalty Scores. The first category includes two
metrics; "total time", representing the total time needed to complete
the exercise, and "economy of motion", which is the travel distance of the instruments during the exercise. Both metrics have
an initial score of fifty (50) points. When the duration for completing the scenario exceeds a predetermined  threshold or the user makes unnecessary movements (thus increasing the total distance), points are deducted.

For the implementation of the scoring system, we used the analytics
framework from the MAGES SDK. We mainly utilized the custom scoring factors that monitor data from objects. For instance, in order to compute the "economy of motions metric",
we summarize the total changes of both position and rotation for each
bone in our Inverse Kinematic (IK) chain. More details about the IK
solver will be discussed in Section \ref{robot control}. Upon exercise completion a detailed breakdown of the score is presented to the user that also includes the penalty deducted points through
a User Interface (see Fig. \ref{fig:analyticsVR}-Top). Furthermore,
MAGES automatically uploads analytics data for each exercise to a web
browser portal (see Fig. \ref{fig:analyticsVR}-Bottom) after finishing a
training scenario. Users can log in using their credentials and gain access to a detailed log of each training session that showcases their analytics.

\section{Implementation Features \& Novelties}
\subsection{Robot Control}\label{robot control}
The primary objective of this section is to present the control
framework designed to guide the robot in accurately tracking the
user's movements in an intuitive way.
The user holds the VR controllers and by moving them, he can translate and rotate the two machine controllers within the virtual reality environment. These machine controllers, which will be referred to as ``masters'', directly control the robotic arms in the operation room.
Each robot arm's end effector (EE) should achieve the
corresponding desired pose (as extracted by the masters) accurately in real-time without significant delay.
To this end, the computational complexity of the code is of pivotal importance when developing such a framework.

First, the 6-DoF pose (position and orientation) of each master is
mapped into the desired pose for the corresponding EE (left and right). The
mathematical expressions in this section will be formulated for one
master and one EE since the left and right arms are considered
equivalent.
To prevent gimbal locks and other issues associated with representing rotations using Euler angles, rotation matrices are employed to express rotations, while transformation matrices are utilized for poses. 
Let $\bm{T}_{W}^{M} \in SE(3)$ be the pose of the master with respect to an inertial frame, or, world frame. $SE(3)$ is the \textit{Special Euclidean} group in three dimensions, while $\bm{T}_{W}^{M}$ is a $4\times4$ homogeneous transformation matrix represents the translation and rotation from world frame $(W)$ to the master frame $(M)$ and it is defined as:  
\begin{equation}
    \bm{T}_{W}^{M} =
\begin{bmatrix}
    \bm{R}_{W}^{M} & \bm{p}\\
    \bm{0} & 1 \\
\end{bmatrix}
\end{equation}
where $\bm{R}_{W}^{M} \in SO(3)$ is the $3\times3$ rotation matrix which belongs to the \textit{Special Orthogonal} group, $\bm{p} \in \mathbb{R}^3$ the translation part, $\bm{p} = \left[ x,y,z \right]^{T}$ and $\bm{0} = \left[0,\;0,\;0 \right]$. 
At each iteration, the orientation of the master is mapped to the EE's frame to extract the desired orientation $(R_{d})$:
\begin{equation}\label{eq:rotation}
    R_{d} = R_{Wt}^{M} \times (R_{W0}^{M})^T \times R_{W0}^{EE}
\end{equation}
where, $R_{Wt}^{M}$ is the current orientation of the master at time $t$, $(R_{W0}^{M})^T$ is the transpose (or inverse) of the initial orientation of the master and, $R_{W0}^{EE}$ is the initial orientation of the EE.

For computing the desired position of the EE, $\bm{p}_d := \left[x_d,\;y_d,\;z_d\right]$, with respect to its own initial position, the following formula is utilized:
\begin{equation}\label{eq:translation}
    \bm{p}_d = \bm{p}_{0}^{EE} + \alpha \cdot (\bm{p}_{t}^{M} - \bm{p}_{0}^{M})
\end{equation}
where $\bm{p}_{0}^{EE}$ is the initial position of the EE,
$\bm{p}_{t}^{M}$ is the current position of the master, $
\bm{p}_{0}^{M}$ is the initial position of the master and finally
$\alpha$ which denotes the translation sensitivity multiplier.
Moreover, $\alpha$ is a hyperparameter that specifies the amount of
change in the position of the EE for a certain displacement of
the master. Given that this implementation is specifically tailored
for surgical systems, enhancing the overall accuracy requires
reducing the movement of the EE to considerably
less than that of the master's displacement $\left(\alpha \ll
1\right)$.

\begin{figure}[t]
    \centering
    \includegraphics[width=\columnwidth]{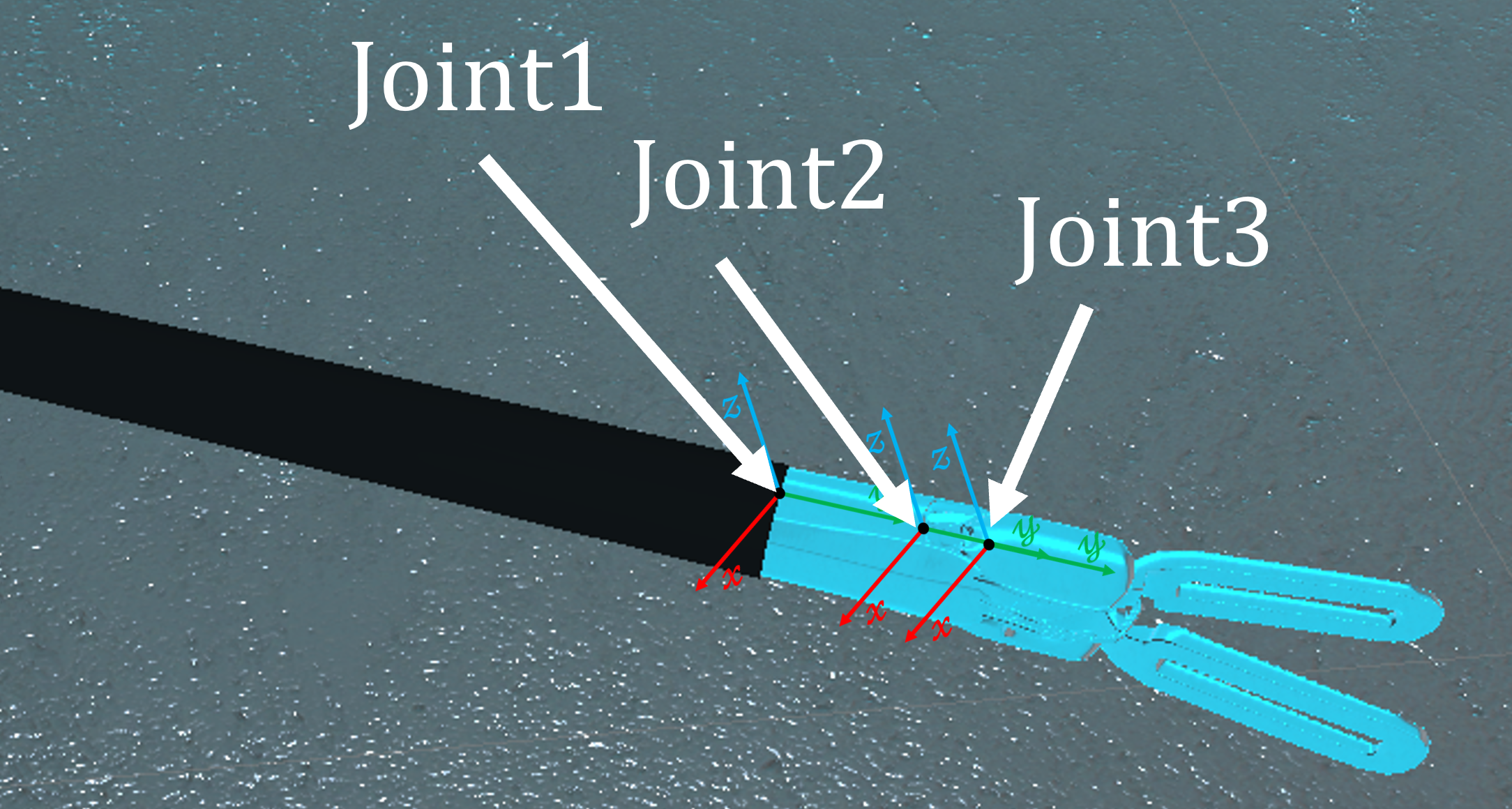}
    \caption{Three revolute wrist joints of the digital-twin robotic arm with the corresponding axes at each center of rotation.}
    \label{fig:joint_conf}
\end{figure}
At every time step, after extracting the desired pose of the EE using equations (\ref{eq:rotation}) and (\ref{eq:translation}), the corresponding control inputs ($u\in \mathbb{R}^6$) that achieve this pose are calculated using Inverse Kinematics (IK). 
To this end, we have developed a numerical-approximation IK solver that is utilized to determine the control inputs for each fully-actuated manipulator as follows. Consider the forward kinematics equation $\bm{x} = \bm{f}\left(\bm{\theta}\right)$, where $\bm{f}: \mathbb{R}^6 \rightarrow \mathbb{R}^6$ is the forward kinematics function, $\bm{x}\in \mathbb{R}^6$ is the pose of the EE  and $\bm{\theta} \in \mathbb{R}^6$ are the joint angles of the robot. The three wrist joint angles are depicted in Fig.~\ref{fig:joint_conf}.  By deploying the \textit{Newton-Raphson} method to solve the equation $\bm{g}(\bm{\theta}_d) = \bm{x}_d - \bm{f}\left(\bm{\theta}_d\right) = 0$, we extract the desired joint angles $\bm{\theta}_d$ that will result the desired pose for the EE $\bm{x}_d$ by using the $1^{st}$ order Taylor expansion~\cite{modern_robotics}:
\begin{equation}
    \bm{\theta}_d = \bm{\theta}_{t} + \bm{J}^{-1}\left(\bm{\theta}_t\right) \left(\bm{x}_d - \bm{x}_t \right),
\end{equation}
where, $\bm{\theta}_{t}$ is the current state of the robot, $\bm{x}_d$ and $\bm{x}_t$ are the desired and current pose of the EE respectively. $\bm{J}^{-1}\left(\bm{\theta}_d\right)$ is the inverse Jacobian matrix that maps changes from the task space into joint space. 
Ultimately, after computing $\bm{\theta}_d$, we apply it to Unity's articulation drive component (\url{https://docs.unity3d.com/Manual/class-ArticulationBody.html}), which utilizes the information to extract the necessary forces for a smooth transition from the current to the desired joint angles.

\subsection{Hand Tracking}
All SRS machinery is typically controlled via hand
movements of the user through a controller-like system, as depicted in
Figure \ref{fig:davinci_surgeon_console}. Similarly, this approach can
be adopted in the corresponding VR digital-twin, utilizing the HMD
controller. However in order to increase the system's portability whilst enhancing the intuitiveness of controlling the robotic arms
\cite{Trimanipulation}, we opt to explore replicating controller
movements through hand tracking technology.  This technology could
enhance the training experience, particularly because the pitcher-like
movement of the actual SRS cannot be replicated with a commercially
standard VR controller. Therefore, hand gestures could be readily
identified using hand tracking.

\begin{figure}
    \centering
    \includegraphics[width=\columnwidth]{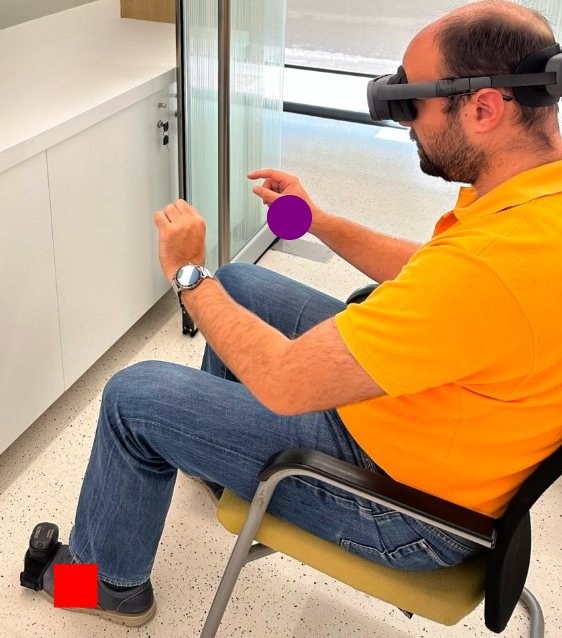}
    \caption{In this photo the user engages in exercises using hand tracking for controlling the robotic arms and HTC Vive Ultimate Trackers to operate the pedals on the surgeon's console. HTC Vive Ultimate Trackers are represented by red squares, while the user's hand, governing the console's masters, is depicted by a purple circle.}
    \label{fig:handtacking_alexandros}
\end{figure}

In VR Isle Academy, we explored hand tracking with Wave SDK on the HTC Vive XR Elite headset and the XR Hands (\url{https://docs.unity3d.com/Packages/com.unity.xr.hands@1.1/manual/index.html}) with the OpenXR (\url{https://www.khronos.org/openxr/}) loader on the Meta Quest 2/3 headsets. We aimed to replicate users' interactions and hand poses on the surgeon's console of an SRS, which led to the integration of hand tracking as can be observed in Fig. \ref{fig:handtacking_alexandros}.

Notably, the Wave SDK doesn't offer a straightforward method for pinch-based
grabbing, in contrast to Unity's XR Hands, which include a built-in
interaction mechanism.

Notably, hand tracking, by relying solely on onboard VR sensors, has limitations. For instance, the user's hands might exceed the Field of View (FOV) of the VR cameras which results in loss of tracking.
Another significant issue is self-occlusion in instances that the user engages in intricate hand poses. This complication led to the VR system
being unable to accurately recognize when the user was pinching,
consequently preventing the forceps from closing in the digital twin.

While we successfully addressed the FOV problem by
incorporating two HTC Vive Wrist Trackers on the user's wrists, the
persistent self occlusion issue prompted. The hand tracking feature reserved solely for specific scenarios where pinch gesture is not required (such as Wrist Articulation). As a result, the pinch-like 
gesture that an SRS user would perform is instead performed via trigger buttons by the VR trainee.

\subsection{Feet Tracking} \label{feetTrackingSeciton}

Our primary objective was to implement a foot tracking solution to
interact with physical pedals within the VR, which in turn would
trigger specific functionalities of the surgical robot. Initially,
HTC Vive Trackers 3.0 were considered, but their reliance on external
lighthouse cameras made them less suitable for our standalone
application and their setup was not as straightforward as desired.
Consequently, we opted for the HTC Vive Ultimate Trackers (\url{https://www.vive.com/us/accessory/vive-ultimate-tracker/}), primarily due
to their compatibility with standalone VR system such as the HTC Vive XR
Elite, and particularly for Android applications. 
HTC Vive Ultimate Trackers are an enhancement of the standard model. They are distinguished
by the inclusion of two 3D cameras, which significantly refine the
mapping of the user's surroundings. This addition allows for a more
nuanced interaction within virtual environments.

The Wave Tracker Manager from the Essence package is utilized in our
application. It contains class references for each tracker, including
the tracker ID. The manager provides two key functionalities:
activating the initial start tracker, triggering the tracker interface
for direct communication when the API starts, and enabling the use of XR
Device to retrieve tracker data from
\texttt{UnityEngine. XR. InputDevice}
(\url{https://docs.unity3d.com/ScriptReference/XR.InputDevice.html}). 
This class is part of Unity's XR input subsystem, managing
input devices in XR applications, defining an XR input device, and
handling input features and haptic feedback. The script
checks for instances where trackers may become stuck, specifically
when they have a valid rotation but no positional data, issuing warnings
to the user. It also verifies tracker connections based on their
specific ID within the application.
Once communication is established between the trackers and the headset, we implement the rotation and position data from the trackers onto the feet of the Full-Body avatar, ensuring an ongoing synchronization with the user's movements.
\subsection{Feet tracker Mini-map}

Mini-map in video games is a small heads-up display (HUD) map which is usually
placed at the corners of the screen to help the player navigate 
inside the virtual world. Usually, the mini-map contains topographical information regarding key objectives and world features.

\begin{figure}
    \centering
    \includegraphics[width=\columnwidth]{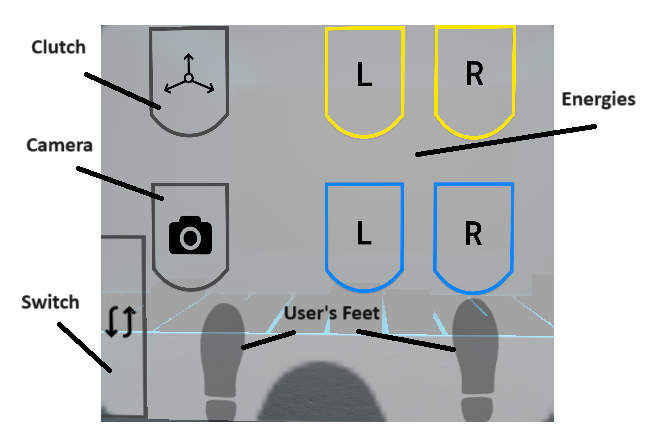}
    \caption{The mini-map of the pedals in the digital twin.}
    \label{fig:vracademy_minimap_2}
\end{figure}

\begin{figure}
    \centering
    \includegraphics[width=\columnwidth]{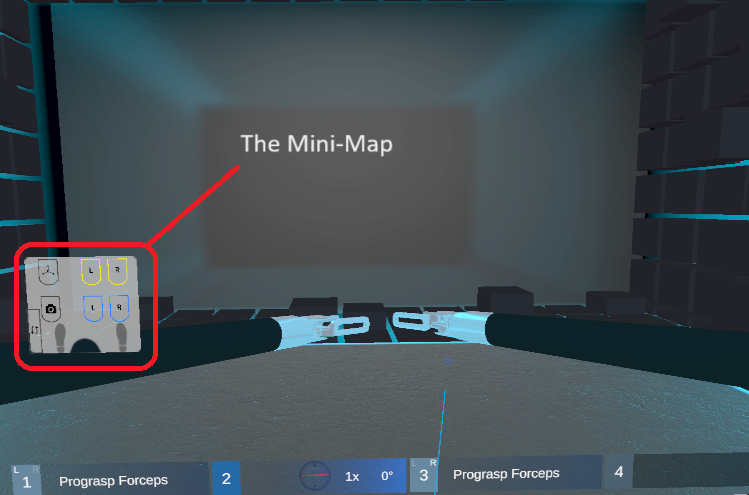}
    \caption{The view from the surgeon's console simulation in the digital twin surgical robotic system VR Isle Academy.}
    \label{fig:vracademy_minimap_1}
\end{figure}

The utilization of feet trackers enables the user to literally press the pedals inside the virtual world. However, such action requires some form of feedback that will notify the user when the pedal has been pressed successfully. However, a physical form of feedback (vibration) is unavailable in HTC Vive Ultimate Trackers. To this end, we propose a visual feedback scheme by utilizing the minimap.
We created a User-Interface (UI) that represents the placement of the
pedals and two UI elements, one for each foot (Fig.
\ref{fig:vracademy_minimap_2}). The mini-map is placed on the left
side of the machine's screen (Fig. \ref{fig:vracademy_minimap_1}).
For the pedals, the user can see the Clutch, Camera, Switch and all
the energy types. These pedals are static, they cannot change position
or orientation. From the other half, the user's feet UIs change
position and scale. when the user moves his feet, the UI will change
position. Then, by pressing one of the buttons, the UI icon of the
particular pedal will turn black and an audible "click" sound will be triggered in
order to inform the user that he has pressed a pedal. Moreover, our approach solves another issue that corresponds to the height of
the user's feet. If the user raises his leg, the feet UI will scale-up, signifying the foot displacement with the respect to the ground.

\section{Results}

To assess the effectiveness of our application and the impact on facilitating the medical training procedure, we conducted experiments involving 4 testers. The test\-ers are medical students that do not possess any prior knowledge on operating an SRS. Each tester played 4 scenarios, repeating them 3 times. The order of scenarios was Wrist Articulation 1, Clutch, Camera 0, Sea Spikes 1, and Roller Coaster 1. Wrist Articulation 1 is the easiest scenario, whereas Roller Coaster 1 is the most challenging. The first three scenarios aim to familiarize the user with the core functionalities of the VR Isle Academy, namely, the control of the robotic arms, the clutch pedal and the camera movement. The remaining two scenarios involve a combination of robotic arms and camera movements, requiring users to execute precise actions.

\begin{figure}
    \centering
    \includegraphics[width=0.9\columnwidth]{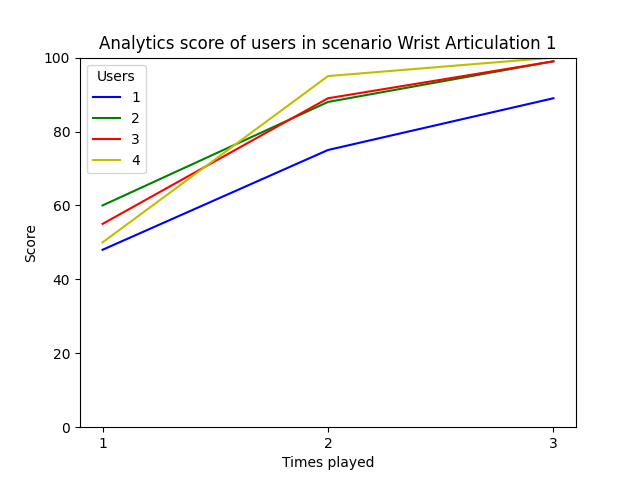}
    \caption{Performance results for the Wrist Articulation 1 scenario across three runs for the four users.}
    \label{fig:wa1}
\end{figure}

In Fig. \ref{fig:wa1}, it is noticeable that the users achieved a higher score as they progress the training scenarios. This scenario is relatively easy, involving the task of learning how to manoeuvre the forceps correctly with minimal wrist movements.

\begin{figure}
    \centering
    \includegraphics[width=0.9\columnwidth]{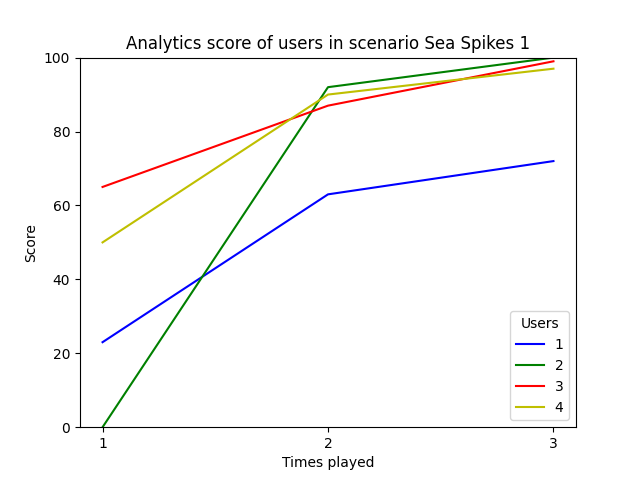}
    \caption{Performance results for the Wrist Articulation 1 scenario across three runs for the four users.}
    \label{fig:ss1}
\end{figure}

In Fig. \ref{fig:ss1}, a distinction is evident between play 1 and play 2. This is attributed to the scenario's combination of correct wrist movement and pedal functionalities, making it challenging to grasp in a single attempt.

The testers found it easy and intuitive to operate the surgeon's console for various training scenarios. The more they played, the easier it became for them to use. The most challenging concept to grasp was the feet trackers. Initially, users reported difficulty localizing their feet and whether they were pressing a pedal.  This might stem from the fact that HTC Vive Ultimate trackers do not provide haptic feedback; hence, the mini-map and sound served as the only source of feedback. Upon familiarization, the users improved at controlling the trackers, getting accustomed to the mini-map.

\begin{figure}
    \centering
    \includegraphics[width=0.9\columnwidth]{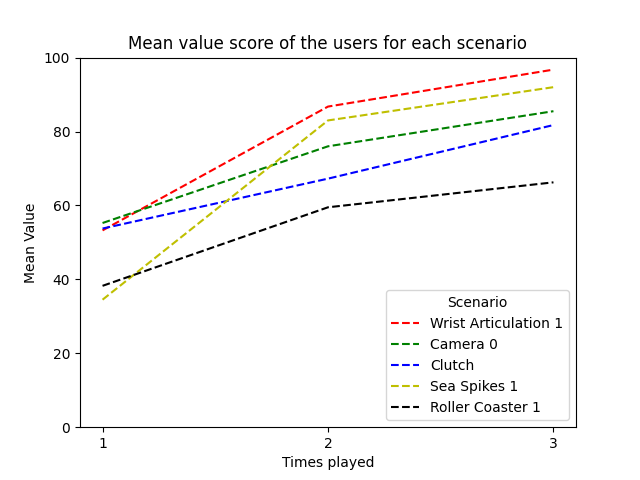}
    \caption{The average score of the four users in each scenario.}
    \label{fig:av_score}
\end{figure}

Even though the testers we selected are aspiring doctors, their background may not be fully aligned with the performance they achieved in the training scenarios. The ability to control a robotic arm should be intuitive and easy regardless of the user's background.

Lastly, users who have actually tried the VR Isle Academy prior to using the real machine have reported an intuitive transition between the two. The required training time with an instructor was significantly reduced and their score was higher than the average.

\section{Conclusion}

VR Isle Academy is a cost-effective solution that enables unsupervised training on operating an advanced SRS. Whether utilizing trackers, VR controllers, or Hand
Tracking, the user enjoys the freedom to train in various settings.
Thus, we minimized the need of using an explicit, bulky device. Also,
the availability of the application is 24/7. Also, there's no need to
schedule a slot or incur additional costs for a teaching service; the
user can independently learn how to operate a Surgical Robotic
System (SRS). 

Moreover, the advancements in VR tracking technology, especially with
the introduction of HTC Vive Trackers and HTC Vive Ultimate Trackers, have
substantially augmented the depth and breadth of virtual reality
applications. These developments have not only heightened the
immersive quality of VR but also expanded its practical applications
across diverse fields. The ongoing evolution of VR technology promises
further enhancements in tracking precision and user engagement.

Even as users strive to familiarize themselves with tracker usage, the
inclusion of a mini-map depicting the position and orientation of the
feet has proven highly beneficial. This approach effectively addresses
the feedback-related challenges by providing visual and auditory cues,
enabling users to orient themselves correctly and press the intended
pedal.

The accurate error detection and comprehensive analytics are pivotal
in such simulations. Surgeons undergoing training in this VR digital
twin will benefit significantly from robust error detection mechanisms
and detailed analytics. The user can check his analytics in real-time,
while using the VR headset, or offline, by accessing a portal page. 

Although we explored the Hand Tracking approach, we found it less
suitable. The ergonomic design of the machine often led to hands going
beyond the FOV of the VR cameras. Additionally,
self-occlusion occurred when adopting unnatural hand poses. This
presented challenges as the application couldn't accurately detect
when the user was "tapping" or "closing" their fingers, impacting the
simulation's fidelity. While we addressed the FOV issue by
incorporating wrist trackers from HTC Vive, the problem of self-occlusion persisted.
Although readily available tools were utilized for developing VR Isle Academy, the novelty lies in their effective integration. The combination of different features and techniques
provides the users with a unique, immersive educational experience and
value.

\section{Future work}

Development will continue, introducing an additional twelve scenarios aimed at training users in utilizing the energy pedals, the switch pedal, and a pedal facilitating the transition between two robotic arms. Additionally, there will be diversification in the types of forceps, incorporating both mono-polar and bi-polar options, enabling the utilization of different energy sources.

Currently, the sole method of employing the trackers
independently is through the Wave SDK. However, there are plans to
ensure compatibility of the Vive Ultimate Trackers with OpenXR. Preparations have been made, having already executed a
port of the VR application to OpenXR for other HMDs like Meta Quest 2 and Meta Quest 3.

Finally, controlled clinical trials involving surgeons will be conducted to assess the fidelity and resemblance between the real and the digital twin SRS. Two groups will be formed. Both groups will use the real SRS while only one of them will have been trained using VR Isle Academy. Subsequently, we will compare the efficiency and time taken by each user, comparing the usage of our application and an SRS versus using only an SRS.

\begin{acks}\label{sec:acks}
Heartfelt gratitude is extended to George Sofianos, Angeliki Karava and Alexandros Sgouridis for their generous provision of a modern SRS system and invaluable help and support throughout the learning journey.
Lastly, special thanks to Pearly Chen and HTC Vive for providing us early access to the HTC Vive Ultimate body trackers.

This work is partially supported by the OMEN-E project (PFSA22-240), that have received 
funding from the Innosuisse Accelerator programme.

\end{acks}

\bibliographystyle{ACM-Reference-Format}
\bibliography{bibliography}

\end{document}